\def\year{2019}\relax
\documentclass[letterpaper]{article} 
\usepackage{aaai19}  
\usepackage{times}  
\usepackage{helvet}  
\usepackage{courier}  
\usepackage{url}  
\usepackage{graphicx}  

\begin{document}
%

\title{Scene Text Detection with Supervised Pyramid Context Network}
\author{
Enze Xie$^{1,3*}$, Yuhang Zang$^{2,3}$\thanks{ indicates equal contribution. This work was done when Enze Xie and Yuhang Zang were interns in Detection Group, Face++, Beijing, China.}, Shuai Shao$^{3}$, Gang Yu$^{3}$, Cong Yao$^{3}$, Guangyao Li$^{1}$\thanks{ Corresponding author.}\\
${^1}$Department of Comuter Science and Technology, Tongji University\\
${^2}$School of Information and Software Engineering,  University of Electronic Science and Technology of China\\
${^3}$Megvii (Face++) Technology Inc.\\
\{xieenze, lgy\}@tongji.edu.cn, yuhangzang@foxmail.com, \{shaoshuai, yugang, yaocong\}@megvii.com
}

\maketitle

\begin{abstract}
Scene text detection methods based on deep learning have achieved remarkable results over the past years.
However, due to the high diversity and complexity of natural scenes, previous state-of-the-art text detection methods may still produce a considerable amount of false positives, when applied to images captured in real-world environments. To tackle this issue, mainly inspired by Mask R-CNN, we propose in this paper an effective model for scene text detection, which is based on Feature Pyramid Network~(FPN) and instance segmentation. We propose a supervised pyramid context network~(SPCNET) to precisely locate text regions while suppressing false positives.

Benefited from the guidance of semantic information and sharing FPN, SPCNET obtains significantly enhanced performance while introducing marginal extra computation. Experiments on standard datasets demonstrate that our SPCNET clearly outperforms start-of-the-art methods. Specifically, it achieves an F-measure of 92.1\% on ICDAR2013, 87.2\% on ICDAR2015, 74.1\% on ICDAR2017 MLT and 82.9\% on Total-Text.
\end{abstract}

\section{Introduction}

Reading text in the wild, as a fundamental task in the field of computer vision, has been widely studied. 
Many applications in the real world rely on accurate text localization, such as license plate recognition, autonomous driving, and document analysis.
Recently, most previous works mainly focus on several challenging issues in natural scene text detection, such as multi-oriented text~\cite{lyu2018multi}, large aspect ratios~\cite{liao2018rotation}, and difficulty in separating adjacent text instances~\cite{deng2018pixellink}. 
However, due to the large differences in foreground text and background objects, as well as the variety of text changes in shape, color, font, orientation and scale, together with extreme illumination and occlusion, there are still many challenges to be addressed for text detection in natural scenes.

\begin{figure}[ht!]
\centering
\includegraphics[scale=0.4]{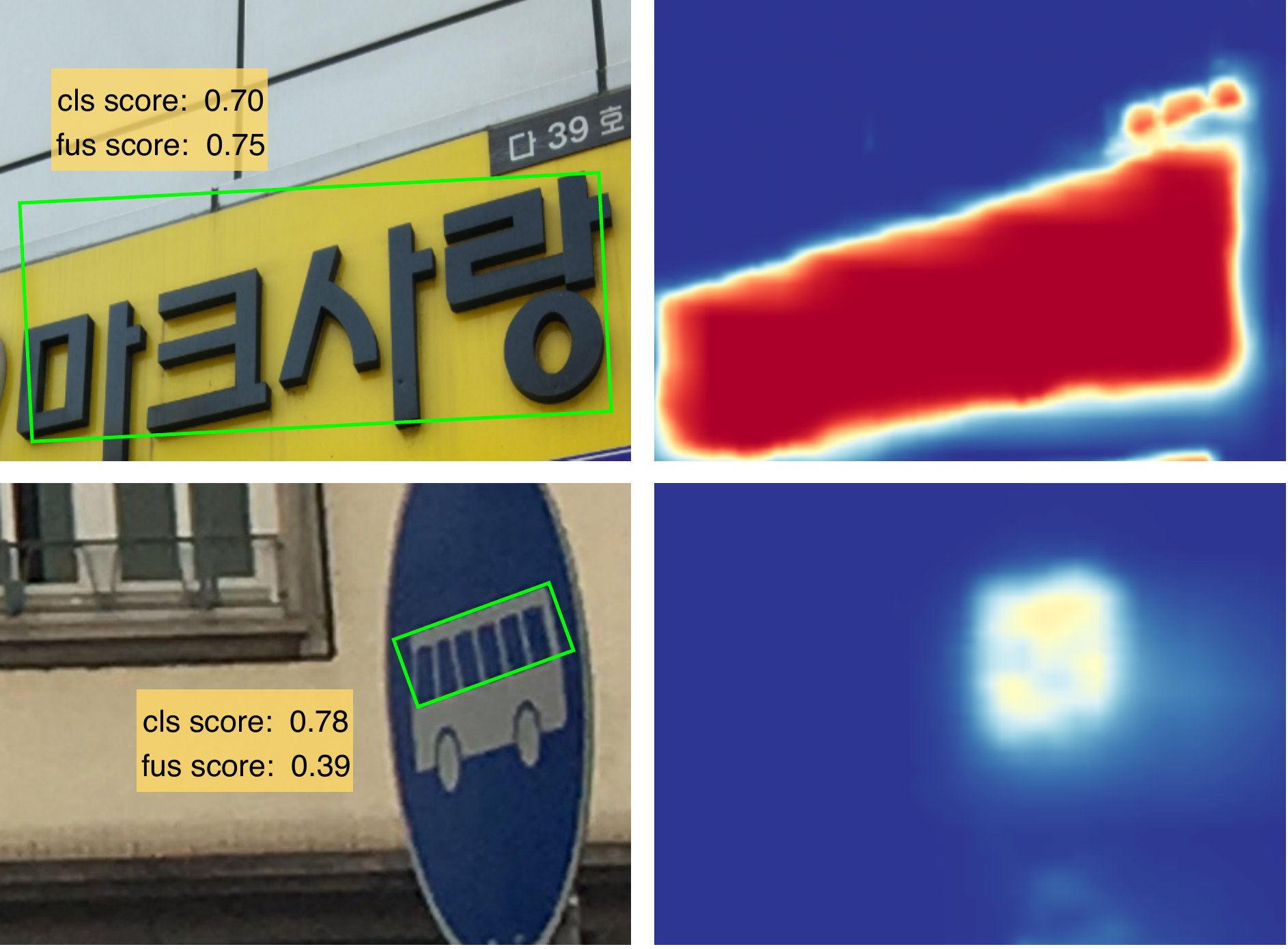}
\caption{\label{visual}{
Visualization of detection results and semantic segmentation feature maps.
Left:The detection result with classification score and fusion score.
The fused score is calculated by Re-Score mechanism. 
Right: The feature map for text segmentation.
}}
\end{figure}

The first challenge is false positives (FP). Some specific scenarios such as autonomous driving require high precision in text detection. 
To the best of our knowledge, little research pays attention to false positive problem in scene text detection. 
Second, flexible locating text in arbitrary shape still remains challenges. 
Text in natural scenes can be in multi-oriented, multi-lingual or curved forms, making network difficult to distinguish FPs.   
Most of the existing methods are specifically designed to detect multi-oriented text and may fall short when handling with curved text. 
TextSnake~\cite{long2018textsnake} uses ordered disks to represent curved text, but it still needs time-consuming and complicated post-processing.

To detect text with various forms, instance segmentation based method is adopted. 
Modern instance segmentation methods, such as Mask R-CNN~\cite{he2017mask}, are usually developed as a multi-task learning problem:
(1) differentiate foreground object proposals from background and assign them with proper class labels. 
(2) perform regression and segmentation on each foreground proposal. 

Nevertheless, simply transfer Mask-RCNN to the text detection scenario is prone to cause some problems, for the following two reasons:
~(1) \textbf{Lack of context information clues}. 
False positives in natural scene tend to be closely related to the surrounding scene. 
For instance, dishes often appear on the table, and fences usually appear in batches.  
However, Mask R-CNN distinguishes object in a single region of interest, which lacks global semantic information guide. Thence, it tends to cause classification errors on some objects who have similar texture information to text without the helping of context information clues. 
~(2) \textbf{Inaccurate classification score}.
The classification scores of Mask R-CNN are easily to be inaccurate when dealing with tilted text. 
Because for tilted text, Mask R-CNN gives classification score rudely based on horizontal proposal, while the background occupies a large proportion. Therefore, when facing tilted text, the classification score of Mask R-CNN tends to be low.

In this paper, we propose a shape robust text detector guided by semantic information. 
Inspired by Mask R-CNN, which can generate shape masks of objects, we use the output of the mask branch to locate the text area. Thus our method is flexible to detect text of arbitrary shapes.

In order to solve the FP problems of lacking context information clues and inaccurate classification score, we design the Text Context module and Re-Score mechanism. 
For Text Context module, we use the semantic segmentation branch to auxiliary guide the detection branch capturing the context information. Through compensating global semantic feature, the network discriminates FPs better.
For Re-Score mechanism, we compensate activation values on segmentation map to classification score to get a fused score. When tackling with tilted text, although the classification score is relatively low, the response on the segmentation map remains strong, leading to an accurate high fused score.
The Re-Score mechanism can further help to reduce FP numbers. This is because the response of FP on segmentation map is intensely weak, causing low fused score. Therefore, FPs with low scores will be more easily filtered out during inference.
The visualization result of the Re-Score mechanism is shown in Fig.~\ref{visual}.

Compared with baseline, the proposed algorithm enhances performance significantly, while adding little computation.
Furthermore, the proposed algorithm achieves an F-measure of 92.1\% on ICDAR2013, 87.2\% on ICDAR2015, 74.1\% on ICDAR2017MLT and 82.9\% on Total-Text,
outperforming previous state-of-the-art algorithms in various kinds of scene text benchmarks~(e.g., horizontal, oriented, multi-lingual and curved).

The contributions of this work are three-fold:
(1) We propose Text Context module and Re-Score mechanism, which can effectively suppress false positives. 
(2) The proposed method can flexibly detect text in various shapes, including horizontal, oriented and curved text.
(3) The proposed algorithm significantly outperforms state-of-the-art methods on several benchmarks containing text instances of different forms.

\section{Related Work}

Scene text detection, as one of the most important problems in computer vision, has been extensively studied. Most of the previous deep learning methods can be roughly divided into two branches: segmentation-based text detection and regression-based text detection. 

Mainstream segmentation-based approaches are inspired by fully convolutional networks~(FCN)~\cite{Long_2015_CVPR}. 
\cite{zhang2016multi} first uses FCN to extract text blocks and detect character candidates from those text blocks with MSER.
\cite{yao2016scene} treats one text region as consisting of three parts:text/non-text, character classes, and character linking orientations, then use them as labels for FCN. 
PixelLink~\cite{deng2018pixellink} performs text/non-text and link prediction on an input image, then adds some post-processing to get text box and filter noise.
PSENET~\cite{li2018shape} finds text kernels and uses progressive scale expansion to position text boundary.
\cite{peng2017large} argues that using large kernel can help boosting semantic segmentation performance.
The main difference between these methods is the generation of different labels for the text. 
Segmentation-based approaches often need time-consuming post-processing steps while obtained performance is still unsatisfying.

General object detection and instance segmentation methods,~e.g., Faster R-CNN~\cite{ren2015faster}, SSD~\cite{liu2016ssd} and FCIS~\cite{li2016fully}, are widely applied to text detection. 
TextBoxes~\cite{liao2017textboxes} modifies anchors and kernels of SSD to detect large-aspect-radio scene text. 
EAST~\cite{zhou2017east} adopts FCN to predict a text score map and a final box for each point in the text region.
RRD~\cite{liao2018rotation} extracts two types of feature for classification and regression respectively for long text line detection.
Based on Faster R-CNN,~\cite{ma2018arbitrary} adds rotation to both anchors and RoIPooling to detect multi-oriented text region.
IncepText~\cite{yang2018inceptext}  uses FCIS to detect multi-oriented text boxes from the perspective of instance segmentation.

However, most of the above methods lack attention to false positives problem in scene text detection, and these methods are often not flexible enough to adapt to arbitrary shapes of text detection. 
In this paper, we devise a pipeline that uses deep supervised semantic information to guide Mask R-CNN finding text area accurately and suppress false positives efficiently.
The model combines instance segmentation with semantic segmentation and allows training in an end-to-end manner.
Moreover, the proposed method can flexibly detect text of arbitrary shape.
Results on several benchmarks show that our method significantly surpasses all previous methods by an obvious gap in performance.

\section{Proposed method}
Our pipeline is composed of two key parts: a Text Context module and a post Re-Score mechanism.
The basis of this pipeline is based on Mask R-CNN.
The text-context module contains two modules: a text attention module and a deep feature fusion module.
This section is organized as follows: 
In Section 3.1, we examine the method for text detection based on Mask R-CNN. 
In Section 3.2, we illustrate the effectiveness of the Text Context module in suppressing false positives. 
In Section 3.3, we show the irrationality of the original scoring method and propose a method of Re-Score to further suppress FPs.
In Section 3.4, we explain the loss function design.

\begin{figure*}[htbp]
\centering
\includegraphics[scale=0.67]{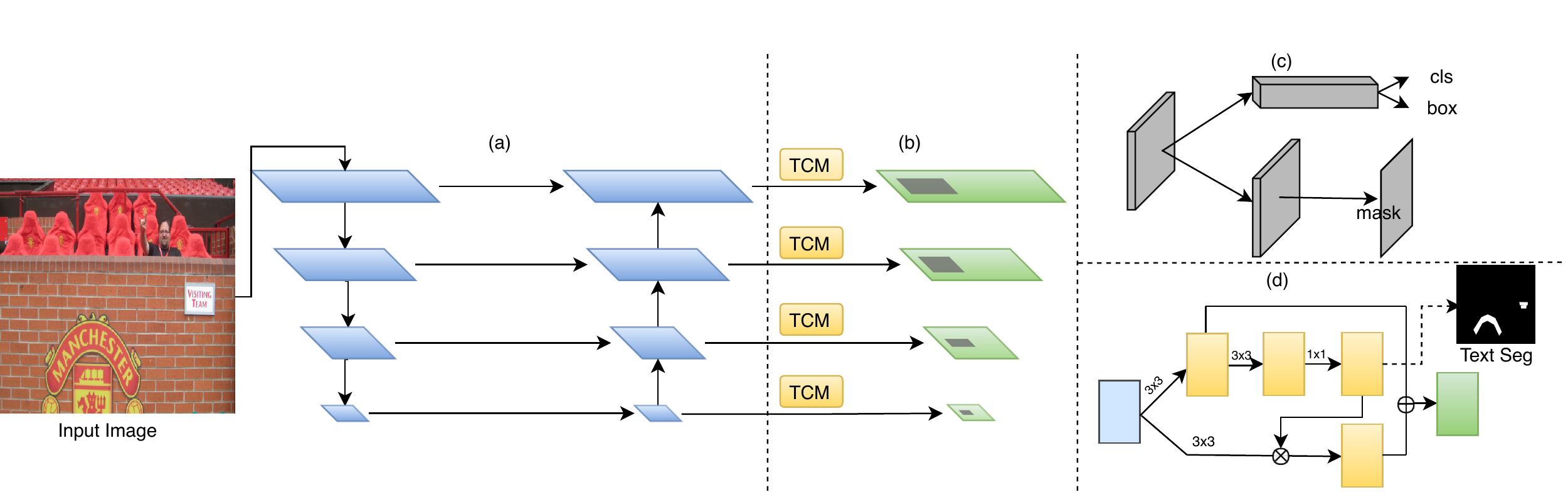}
\caption{\label{network}{The architecture of our method. (a) The Feature Pyramid Network~(FPN) architecture. (b) Pyramid Feature fusion via TCM. (c) Mask R-CNN branch for text classification, bounding box regression and instance segmentation. (d) The proposed Text-Context Module(TCM). Dotted line indicates the text semantic segmentation branch. The text segmentation map is upsampled to the input image size and calculates the loss with Ground Truth. 
}}
\end{figure*}

\begin{figure}[htbp]
\centering
\includegraphics[scale=0.6]{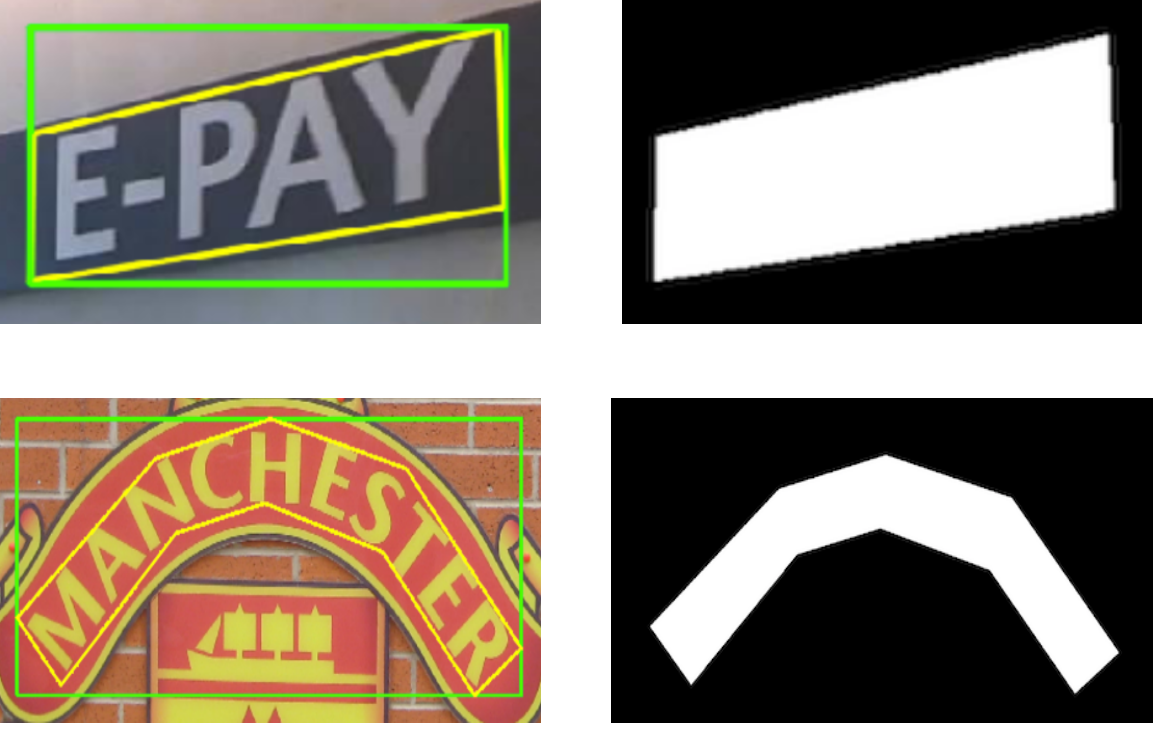}
\caption{\label{label}{Ground truth. Left: Image sample with green bounding box and yellow polygon. Right: Corresponding binary text segmentation map.
}}
\end{figure}

\subsection{Mask R-CNN} 
\textbf{Why Mask R-CNN?} 
Mask R-CNN is the state of the art in instance segmentation.
 Most of the winners in MS COCO instance segmentation challenge are based on Mask R-CNN.
A recent work \cite{lyu2018mask} also uses Mask R-CNN for end-to-end text detection and recognition.
 Hence Mask R-CNN makes a strong baseline to compare against.

\subsubsection{Label Generation} 
The ground truth of text instance is exemplified in Fig.~\ref{label}. Different from common instance segmentation datasets, pixel-level text/non-text annotations are not provided. We treat the pixels in the polygon as text, and the pixels outside the polygon as non-text, then we get an instance of the text area. The minimum bounding horizontal rectangle of the polygon will be treated as a bounding box. We generate the global binary map in the same way as the instance generation.
 
 \subsubsection{Mask R-CNN architecture}
The overall architecture of our proposed method is presented in Fig.~\ref{network}. Our network is composed of five parts: feature pyramid network~(FPN), region proposal network(RPN), R-CNN branch, mask prediction branch and global text segmentation prediction branch.
Feature Pyramid Network~(FPN) is a feature fusion structure widely used in current mainstream detection models. 
FPN uses a top-down architecture with lateral connections to build an in-network feature pyramid from a single-scale input.
Region Proposal Network~(RPN) generates bounding boxes likely to contain an object as proposals.
Through Roi-Align, all proposals are resized to 7$\times$7 for R-CNN branch and 14$\times$14 for mask prediction branch.
The global text segmentation branch acts on each stage of the FPN to generate a semantic segmentation map of the text. 

 \subsection{Text Context Module}
Suppressing false positives is a challenging issue for general object detection and text detection. 
In natural scenes, some regular objects, such as discs, fences, etc., are easily detected as text by the detection network.
Mask R-CNN uses region of interests~(ROIs) to classify whether the proposal is text or background.
However, the text region classification is performed with features extracted from only one region of interest.
Since false positives in natural scenes often do not appear unexpectedly, such as plates are more likely appear on the table, introducing contextual information helps the network extract more discriminative features and accurately classify proposals.
Our Text Context Module~(TCM) is composed of two sub-modules:
Pyramid Attention Module~(PAM) and Pyramid Fusion Module~(PFM).
The feature maps are feed to TCM, which produces text segmentation as output.

 \subsubsection{Pyramid Attention Module}
 Our pyramid attention module is inspired by SSTD~\cite{he2017single}. We additionally add a global text segmentation branch after FPN from stage2 to stage5.
It generates a saliency map of pixel-level text/non-text regions for each FPN layer. 
The attention module and the fusion module share a branch, named text context module, including two 3$\times$3 convolutional layers and one 1$\times$1 convolutional layer.
The output saliency map includes two channels, which means text/non-text map. 
We enhance the saliency map and use it to activate the text area on the feature map.
Specifically, take stage2 as an example, giving an input sample of 512$\times$512, the feature map \begin{math}
S_2 \in R^{128\times128\times256}
\end{math}. 
The generation of saliency map is as follows:

\begin{equation}
\label{gongshi1}
map = Text\_Context\_Module(S_{2})
\end{equation}
\begin{equation}
\label{gongshi2}
saliency\_map = e^{Softmax(map)}
\end{equation}

where Text Context module generates the saliency map with 2 channels. Then after the channel-wise softmax, we obtain the text saliency map. Through the Exponential activation, the saliency map is enhanced, that is, the response gap in text/non-text areas becomes larger.
The saliency map will act on the feature map as follows:

\begin{equation}
\label{gongshi3}
saliency\_map^* = Broadcast(saliency\_map)
\end{equation}
\begin{equation}
\label{gongshi4}
S_2^* = saliency\_map^* \odot S_2
\end{equation}
where saliency\_map is broadcast to the same 256 channel as $S_2$, 
and ``$\odot$'' represents the pixel-by-pixel multiplication of the two maps $S_2$ and $saliency\_map^*$.

\subsubsection{Pyramid Fusion Module}
Next we introduce the pyramid fusion module.
The PFM combines detection feature with the deep supervised semantic feature, makes the network more discriminative to distinguish text from non-text.
Specifically, semantic segmentation examines text from the perspective of a single pixel and determines the text region by combining the information of surrounding pixels, and the detection classifies the text region by ROIs.
There is a natural complementary relationship between the two branches.

After first 3$\times$3 convolutional layers of Text Context module, we get the feature map(GTF) of global text segmentation. 
These features capture complementary information like context, semantic segmentation of background and of text.
Both computer vision~\cite{divvala2009empirical} and cognitive psychology~\cite{oliva2007role} research show that identifying the local surrounding of an object helps to better identify itself. 
This is because the category of object are often correlated with surrounding stuff, e.g. discs often appear on the table. 
Although there is only textual annotation information, this encoding method allows the network to implicitly learn more discriminative semantic information.
Introducing it into the original feature map makes Mask R-CNN performing stronger on the classification task.
The specific details are as follows:

\begin{equation}
\label{gongshi5}
GTF = Conv_{3\times3}(S_2)
\end{equation}
\begin{equation}
\label{gongshi6}
\hat{S_2} = S_2^* + GTF
\end{equation}
where the $Conv_{3\times3}$ is the first Conv layer in Text Context module and GTF represent global text feature. Then ``$+$'' represents element-wise addition operation.

\begin{figure}[htp]
\centering
\includegraphics[scale=0.55]{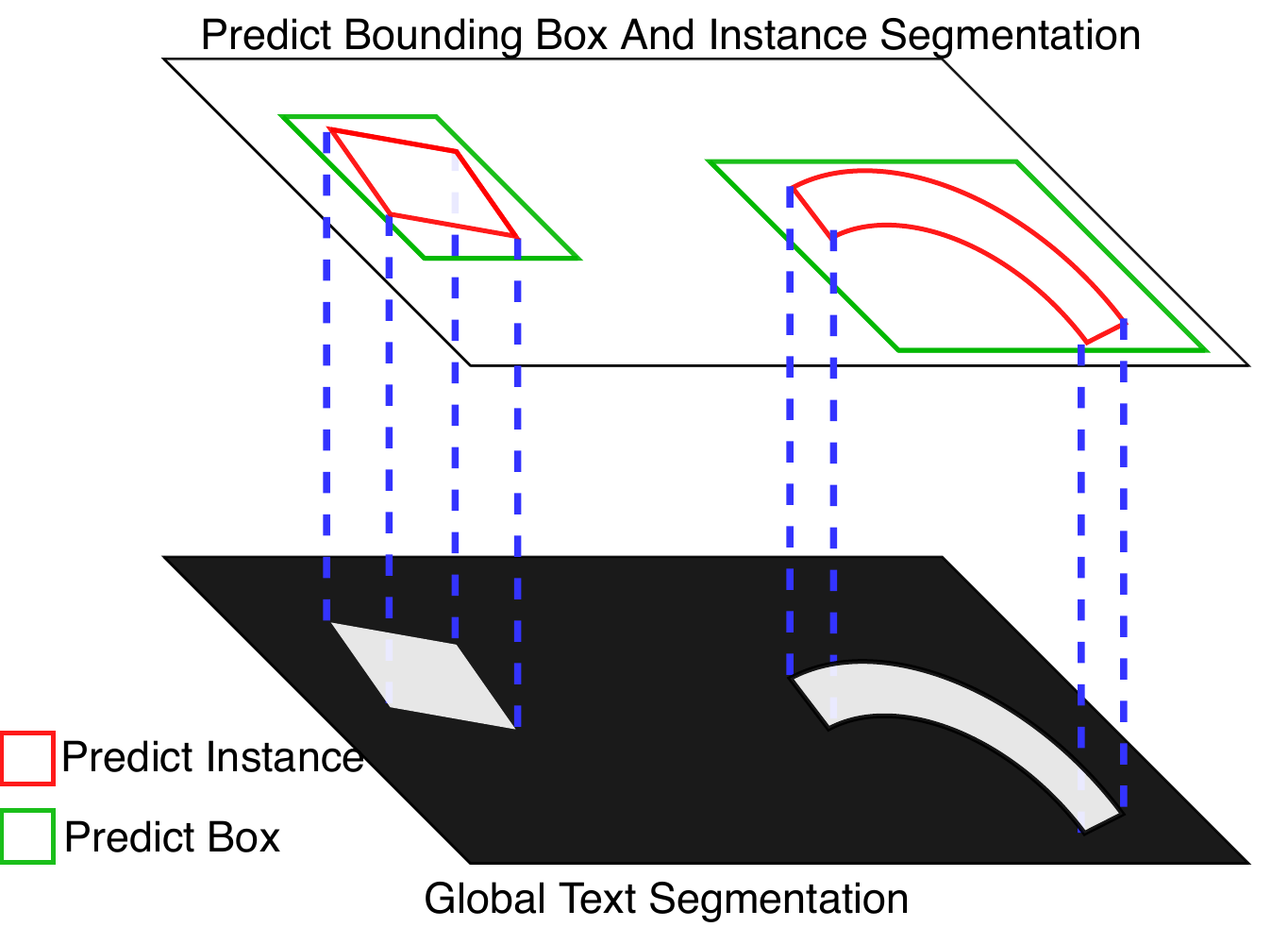}
\caption{\label{rescore}{Overview of the Re-Score Mechanism. Upon:The predicted text boxes and instances of the input images; Bottom:The global text segmentation map output from TCM. For each text instance, we project them onto the segmentation map and calculate the activation value of the projected area. }}
\end{figure}

 \subsection{Re-Score Mechanism}
For standard Mask R-CNN inference processing, the predicted top-K(e.g., 1000) bounding boxes are sorted by the classification confidence, then after standard NMS processing, Up to top-M(e.g.,  300) bounding boxes with highest classification confidence are retained. These bounding boxes are feed to Mask R-CNN as proposals to generate predicted text instance maps. 
This method treats one horizontal bounding box's classification confidence as the score, then artificially sets a threshold to filter out background boxes.
However, this method will filter out some true positives with low scores, because if a horizontal bounding box encloses a titled text instance, it also accompanies a lot of background information. 
At the same time, some FPs with relatively high confidence will be retained.

We re-assign scores for each text instance. The visualization diagram is shown in Fig.~\ref{rescore}. The fused score of text instance is composed of two parts: classification score(CS) and instance score(IS).
Formally, the fused score for the $i$th proposal, given the predicted 2-class scores CS = \{$s_{i0}^{cs}$, $s_{i1}^{cs}$\} and IS = \{$s_{i0}^{is}$, $s_{i1}^{is}$\} is computed via the following softmax function:
\begin{equation}
\label{gongshi7}
s_{i} = \frac{e^{(s_{i1}^{cs}+s_{i1}^{is})}}   {e^{(s_{i1}^{cs}+s_{i1}^{is})} +e^{(s_{i0}^{cs}+s_{i0}^{is})}}
\end{equation}
where CS is directly obtained by Mask R-CNN classification branch, and IS is the activation value of the text instance on the global text segmentation map.
In details, for each text instance, it is projected onto text segmentation map, containing $P_i$ = $\{p_i^1,p_i^2...p_i^n\}$, and the mean of $p_i$ in the text instance area is calculated:
\begin{equation}
\label{gongshi8}
s_{i1}^{cs} = \frac{\sum_{j} p_i^j }{N} 
\end{equation}
where $P_i$ is the set of the pixels' value of $i$th text instance on text segmentation map.
The fused score combines the classification score with the instance score,
which can reduce the FP confidence effectively, because FP instances tend to have weaker response than text on the segmentation map.
This mechanism is also more friendly for titled text, because the titled text instance also has a strong response on the segmentation map, high instance score will compensate for low classification score.

 \subsection{Loss Function Design}
 Similar to Mask R-CNN, our network includes multi-task. 
 Following the loss function design of Mask R-CNN, we additionally add a global text segmentation loss based on it.
 The loss expression is as follows:
 \begin{equation}
\label{gongshi9}
L = L_{rpn} + \lambda_1 \cdot L_{cls} +  \lambda_2 \cdot L_{box} + \lambda_3 \cdot L_{mask} + \lambda_4 \cdot L_{gts}
\end{equation}
where  $L_{rpn}$,  $L_{cls}$,  $L_{box}$ and $L_{mask} $ are the standard loss in Mask R-CNN. 
The $L_{gts}$  is used to optimize global text segmentation, defined as :
\begin{equation}
\label{gongshi10}
L_{gts} = \frac{1}{N}  \sum_{i} {-}\log(\frac{e^{p_i}}{\sum_j {e^{p_j}}})
\end{equation}
The $L_{gts}$ is Softmax loss, where $p$ is the output prediction of the network.

Multitask learning is the process of learning useful representations of multiple complementary tasks from the same input, and has been found to improve the performance of both tasks.
This method enables the network to learn text detection and global text segmentation by end-to-end joint training, allowing gradients from two tasks to influence shared feature maps.

\begin{figure*}[ht!]
\centering
\includegraphics[scale=0.34]{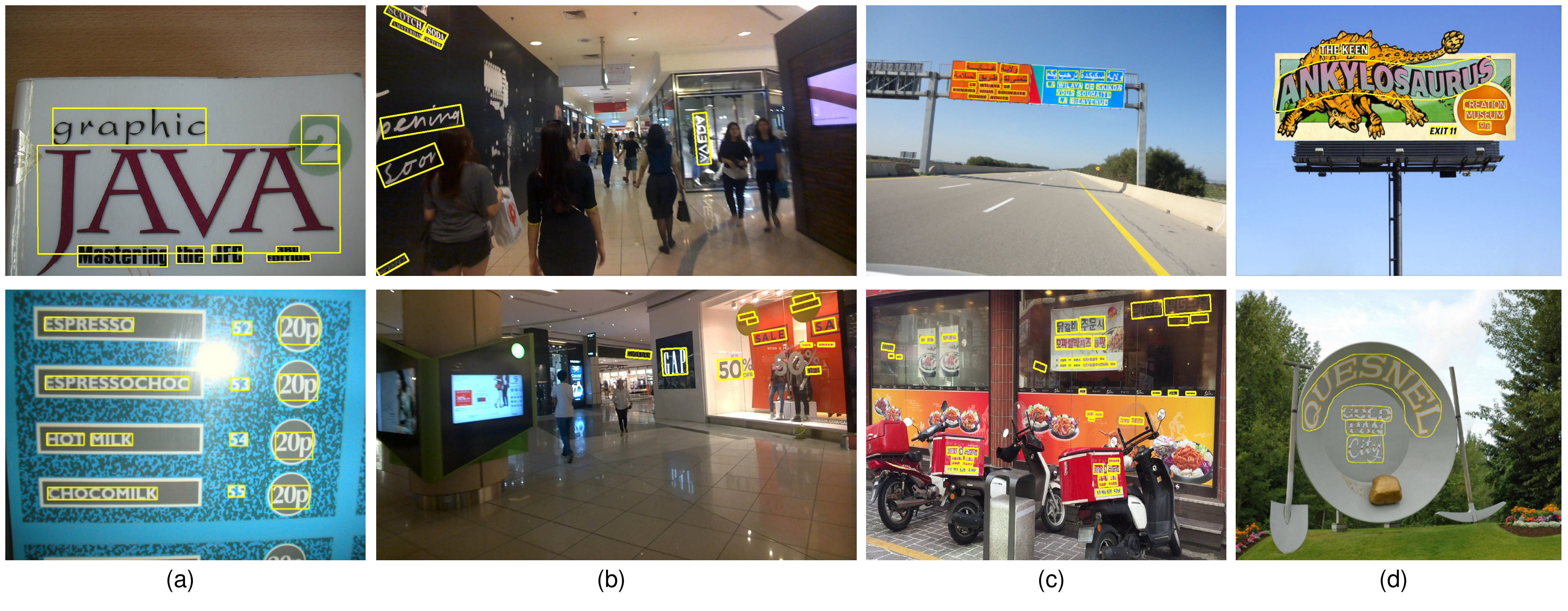}
\caption{\label{result}{Qualitative results of the proposed algorithm. (a) ICDAR2013. (b) ICDAR2015. (c)ICDAR2017. (d) Total-Text.
}}
\end{figure*}

\section{Experiments}
We evaluate our approach on four standard benchmarks: ICDAR2013, ICDAR2015, ICDAR2017 MLT, Total-Text, and compare with other state-of-the-art methods.
\subsection{Datasets}
The datasets used for the experiments in this paper are briefly introduced below:

\textbf{SynthText}~\cite{gupta2016synthetic} is a synthetically generated dataset composed of 800000 synthetic images. We use the dataset with word-level labels to pre-train our model.

\textbf{ICDAR2017 MLT}~\cite{nayef2017icdar2017} is a dataset focuses on multi-oriented, multi-scripting, and multi-lingual aspects of scene text. It consists of 7200 training images, 1800 validation images, and 9000 test images. Image annotations are labeled as word-level quadrangles. We use both training set and validation set to train our model.

\textbf{ICDAR2015}~\cite{karatzas2015icdar} is a dataset proposed for incidental scene text detection. There are 1000 training images and 500 tests images with annotations labeled as word-level quadrangles.

\textbf{ICDAR2013}~\cite{karatzas2013icdar} is a dataset points at horizontal text in the scene. It contains 229 training images and 233 testing images with only horizontal texts. 

\textbf{Total-Text}~\cite{ch2017total} is a newly-released benchmark for curved text detection. 
The dataset is split into training and testing sets with 1255 and 300 images, respectively.

\subsection{Implementation Details}
\textbf{Training} We set hyper-parameters mainly following Mask R-CNN. Our base-model is ResNet50 and pre-trained on ImageNet. All new layers are initialized with a zero-mean Gaussian distribution with standard deviation 0.001. We use Adam as optimizer with batch size 16, momentum 0.9 and weight decay 1e-4 in training. Similar to~\cite{yu2018bisenet}, we apply the ``poly'' learning rate strategy in which the initial rate is multiplied by $(1 - \frac{iter}{max\_iter})^{power}$ each iteration with power 0.9. The initial learning rate is 
2$\times$10$^{-3}$ for all experiments. We first adopted the warmup strategy in~\cite{peng2017megdet}, then we found without warmup the net can still convergence fast. The network only takes 6h and 1h to complete training when use 8 GPUs. The aspect ratios of anchors are set to $1/5, 1/2, 1, 2, 5$ for all experiments.

\textbf{Data Augmentation} 
We follow the data augmentation strategy of Mask R-CNN. Short edges of the images are randomly resized to three scales (640, 720, 800). Then each image is randomly flipped with a probability of 0.5.

\textbf{Post Processing}
Our post processing is simple. We re-score all text instances, then find the minimum bounding rectangle for each text instance.  Finally a polygon NMS is utilized to suppress redundant boxes.
Methods like $minAreaRect$ in OpenCV can be applied to obtain the bounding boxes of text instances as the final detection result.

\subsection{Ablation Study}
To verify the effectiveness of our approach, we do a series of comparative experiments on the ICDAR2017 MLT validation set. 
These experiments mainly focus on evaluating two essential methods in our model: Text Context module(TCM) and Re-Score mechanism(RS).
Table 1 summarizes the results of our models with different settings on ICDAR2017 MLT.

\begin{table}[htb]
\center
\begin{tabular}{|c|c|c|c|}
 \hline
Method & Recall & Precision & F-measure\\
 \hline
Baseline & 			73.4 & 76.2 & 74.7 \\
 \hline
Ours+TCM &			73.4 &80.3 &76.8 \\
\hline
Ours+TCM+RS &		73.4 &84.2 &\textbf{78.5} \\
\hline
\end{tabular}
\caption{\label{ablation}Effectiveness of several modules on ICDAR2017 MLT incidental scene text location task.}
\end{table}

False positive problems often appear in complex natural scenes. 
The MLT dataset is composed of complete scene images which come from 9 languages.
According to our statistics, the smallest text box size on the MLT is less than 20 pixels, and the largest is more than 3000 pixels. So size range of the text box is very different.
 To the best of our knowledge, it is the most challenging public scene text benchmark, hence the experiment results in MLT are convincing. 
The detailed comparison is given in the following.

\textbf{Baseline}
Mask R-CNN architecture without Text Context Module and Re-Score Mechanism.

\textbf{Text Context Module}
Compared with baseline, the Text Context module achieves an improvement of 4.1 percents on precision while keeping the recall  identical.
This implies that the TCM helps network extract more discriminative features of text/non-text and reduced the number of FPs.

\textbf{Re-Score Mechanism}
In the post-processing stage, we use our proposed re-score mechanism to re-rank the scores of all text instances during inference.
Table \ref{ablation} shows our re-score mechanism can further improve precision of 3.9\% based on TCM.
This brings in total 3.8\% F-measure of revenue compared with baseline.
The experimental result proves that the Re-Score mechanism can further suppress FPs with weakly response on the global text segmentation in post-processing stage.

\subsection{Results on Scene Text Benchmarks}
\subsubsection{Detecting MultiLingual Text } 
We first pretrain the proposed network on SynthText for one epoch then fine-tuned on MLT 9000 train and val images for 40 epochs. 
With single scale of 848(short edge), our proposed method achieves an F-measure of 70.0\%, outperforming state of the art methods over 3\%.
 Since there are many small words on the MLT, we apply a simple multi-scale test method with scale $\in [720,1920]$. By merging the results of two scales, the F-measure is 74.1\%, which outperforms all competing methods by at least 1.7\%.  
To our best knowledge, this is the best reported result in literature.
The result is shown in Table \ref{ic17}.

\begin{table}[htb]
\center
\scalebox{0.85}{
\begin{tabular}{|p{4cm}|c|c|c|}
 \hline
Method & Recall & Precision & F-measure\\
 \hline
TH-DL\cite{nayef2017icdar2017}                        			& 34.8 & 67.8 & 46.0 \\
 \hline	
SARI FDU RRPN V1\cite{nayef2017icdar2017}	& 55.5 & 71.2 & 62.4\\
 \hline
Sensetime OCR\cite{nayef2017icdar2017}      	&69.4 &56.9 &62.6 \\
\hline
SCUT DLVClab1\cite{nayef2017icdar2017}        	&54.5 &80.3 &65.0 \\
\hline
Lyu \it{et al.}\cite{lyu2018multi}             				 &55.6 &83.8 &66.8 \\
\hline
Lyu \it{et al.}$^*$\cite{lyu2018multi}       				&70.6 &74.3 &72.4 \\
\hline
Baseline                          							&62.2 &69.2 &65.5 \\
\hline
Ours                          								&66.9 &73.4 &70.0 \\
\hline
Ours$^*$                  									&68.6 &80.6 &\textbf{74.1} \\
\hline
\end{tabular}}
\caption{\label{ic17}Effectiveness of several methods on ICDAR2017 MLT incidental scene text location task. $^*$ means multi scale test.}
\end{table}

\subsubsection{Detecting Oriented Text }
\begin{figure}[ht!]
\centering
\includegraphics[scale=0.7]{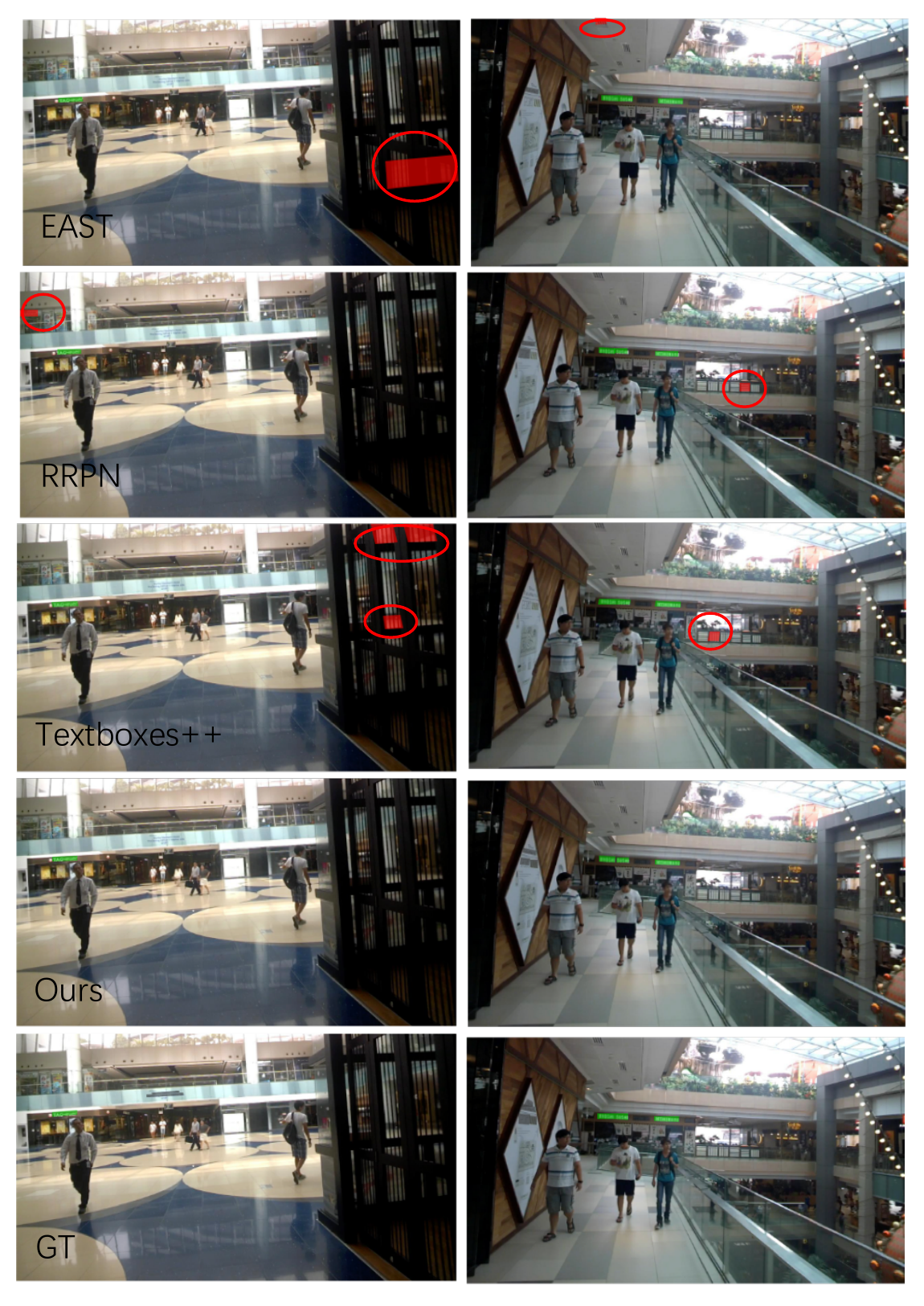}
\caption{\label{FP}{Qualitative detection results of EAST, RRPN\cite{ma2018arbitrary}, TextBoxes++\cite{liao2018textboxes++} and our method. The green and red regions represent true positive and false positive results respectively. Visualizations are captured from the ICDAR official online evaluation
system (http://rrc.cvc.uab.es/?ch=4\&com=evaluation\&task=1).
}}
\end{figure}

On ICDAR2015, we use pretrained model from MLT and fine-tune another 40 epochs.
The comparison with the state of the art results on ICDAR2015 dataset is given in Table \ref{ic15}.
All setting are same as MLT except we only use single scale test.
Experimental results show the results of our method surpasses the state of the art results by more than 1.5\% percents with single scale setting.
Moreover, Fig. \ref{FP} shows our methods can suppress false positives effectively compared with prior arts.

\subsubsection{Detecting Horizontal Text }
On ICDAR2013, the proposed model is pre-trained from MLT and fine-tune on 299 training images for another 40 epochs.
All test settings are the same as ICDAR2015. 
Although our method is specifically designed for text detection of arbitrary shapes, our method also shows superiority in horizontal text detection compared to prior arts. Table \ref{ic13} shows the experiment results of different methods. 
Similarly, in ICDAR2013 dataset, our approach achieves the state of the art result at 92.1\%, experiments prove the effectiveness of our method.

\begin{table}[htb]
\center
\scalebox{0.83}{
\begin{tabular}{|p{4cm}|c|c|c|}
 \hline
Method & Recall & Precision & F-measure\\
 \hline
CTPN\cite{tian2016detecting}                 				& 51.6 & 74.2 & 60.9 \\
 \hline
SegLink~\cite{shi2017detecting}		& 76.8 & 73.1 & 75.0\\
 \hline
  MCN\cite{liu2018learning}               	 				&72.0 &80.0 &76.0 \\
 \hline
 SSTD\cite{he2017single}								&73.0 & 80.0 & 77.0 \\
\hline
 WordSup$^*$\cite{hu2017wordsup}						&77.0 &79.3 &78.2 \\
\hline
EAST$^*$\cite{zhou2017east}                  			&78.3 &83.3 &80.7 \\
 \hline
 Lyu \it{et al.}\cite{lyu2018multi}    			&70.7 &94.1 &80.7 \\
 \hline
 DeepReg\cite{he2017deep}          				&80.0 &82.0 &81.0 \\
 \hline
  RRD$^*$\cite{liao2018rotation}                    			&80.0 &88.0 &83.8 \\
 \hline
  TextSnake\cite{long2018textsnake}      				&80.4 &84.9 &82.6 \\
 \hline
 PixelLink\cite{deng2018pixellink}          			&82.0 &85.5 &83.7 \\
 \hline
 FTSN\cite{dai2017fused}                        		&80.0 &88.6 &84.1 \\
 \hline
 IncepText\cite{yang2018inceptext}         				&80.6 &90.5 &85.3 \\
 \hline
 Baseline                        				&83.8 &87.4 &85.5 \\
 \hline
 Ours                        					&85.8 &88.7 &\textbf{87.2} \\
 \hline
\end{tabular}}
\caption{\label{ic15}Effectiveness of several methods on ICDAR2015. $^*$ means multi scale test.}
\end{table}

\begin{table}[htb]
\center
\scalebox{0.83}{
\begin{tabular}{|p{4cm}|c|c|c|}
 \hline
Method & Recall & Precision & F-measure\\
 \hline
 CTPN\cite{tian2016detecting}                         			& 83.0 & 83.0 & 88.0 \\
 \hline
TextBoxes\cite{liao2017textboxes}                         	& 74.0 & 88.0 & 81.0 \\
 \hline
SegLink~\cite{shi2017detecting}			 	& 83.0 & 87.7 & 85.3\\
 \hline
  MCN\cite{liu2018learning}                        				&87.0 &88.0 &88.0 \\
 \hline
 SSTD\cite{he2017single}		        	 				&86.0 & 89.0 & 88.0 \\
\hline
 WordSup$^*$\cite{hu2017wordsup}		        			&88.0 &93.0 &90.0 \\
\hline
 Lyu \it{et al.}\cite{lyu2018multi}                 	&79.4 &93.3 &85.8 \\
 \hline
 DeepReg\cite{he2017deep}                        		&81.0 &92.0 &86.0 \\
 \hline
  RRD\cite{liao2018rotation}                       				&75.0 &88.0 &81.0 \\
 \hline
 PixelLink$^*$\cite{deng2018pixellink}                    	&87.5 &88.6 &88.1 \\
 \hline
  Baseline                        					&88.1  &91.0  &89.6  \\
 \hline
 Ours                        						&90.5  &93.8  &\textbf{92.1}  \\
 \hline
\end{tabular}}
\caption{\label{ic13}Effectiveness of several methods on ICDAR2013. $^*$ means multi scale test.}
\end{table}

\begin{table}[ht!]
\center
\scalebox{0.85}{
\begin{tabular}{|p{4cm}|c|c|c|}
 \hline
Method & Recall & Precision & F-measure\\
 \hline
SegLink~\cite{shi2017detecting}                      & 23.8 & 30.3 & 26.7 \\
 \hline
 EAST\cite{zhou2017east}                         						& 36.2 & 50.0 & 42.0 \\
 \hline
DeconvNet~\cite{ch2017total}                   & 40.0 & 33.0 & 36.0 \\
 \hline
  TextSnake\cite{long2018textsnake}                         			& 74.5 & 82.7 & 78.4 \\
 \hline
  FTSN\cite{dai2017fused}                         						& 78.0 & 84.7 & 81.3 \\
 \hline
 Baseline                        								&80.5  &81.5  &81.0  \\
 \hline
 Ours                        							&82.8  &83.0  &\textbf{82.9}  \\
 \hline
\end{tabular}}
\caption{\label{total}Effectiveness of several methods on Total-Text dataset. Note that EAST and SegLink were not fine-tuned on Total-Text. Therefore their results are included only for reference.}
\end{table}

\subsubsection{Detecting Curved Text }
We evaluate the ability of our model to detect curved text on Total-Text dataset. 
Similar to the above training methods, we use the MLT pretrained weights to initialization model and fine-tune on Total-Text for 40 epochs.
All test settings are the same as ICDAR2015 and ICDAR2013. 
Our method is shape robust for text detection. 
The proposed method can be flexibly applied to different types of scene text detection datasets without special modifications.
 Experimental results show that our method surpasses prior art methods. 
 The detail results are shown in Table \ref{total}.
 Note that the results of SegLink and EAST are referenced from TextSnake.

\section{Conclusion}

In this work, we have presented a shape robust text detector that can detect text with arbitrary shapes. It is an end-to-end trainable framework with semantic segmentation guidance. We effectively alleviate the false positive problem via introducing context semantic information and re-score mechanism for all predicted text instances.
By sharing convolutional features, the text segmentation branch is nearly cost-free. The results on different scene text benchmarks demonstrate the effectiveness and generalization of our approach.

In the future, we are interested in multiple directions as below:
(1) We will attempt to integrate the Re-Score mechanism into the network in an end-to-end manner.
(2) We are interested in exploring our method on other multi-oriented or curved object detection task, such as an aerial scene. 
(3) We will investigate more efficient fast text detection networks that running on mobile phones. 

\section{Acknowlegement}
This work was supported by the National Natural Science Foundation of China under Grant number 61771346. Special thanks Mr. Mengxiao Lin in Megvii base-model group for all his kindness and great
help to us. 

\bibliographystyle{aaai}
\bibliography{ref/ref} 
\end{document}